\documentclass{article}

\usepackage[final,nonatbib]{mlsafety_neurips_2022}

\usepackage[numbers,sort&compress]{natbib}

\usepackage[utf8]{inputenc} 
\usepackage[T1]{fontenc}    
\usepackage{hyperref}       
\hypersetup{ colorlinks = true, citecolor = blue, urlcolor = blue}
\usepackage{url}            
\usepackage{booktabs}       
\usepackage{amsfonts}       
\usepackage{nicefrac}       
\usepackage{microtype}      
\usepackage{xcolor}         
\usepackage{amsmath}        
\usepackage{enumitem}
\usepackage{graphicx}

\title{Measuring Reliability of Large Language Models through Semantic Consistency}

\author{%
  Harsh Raj \\
  Delhi Technological University \\
  Delhi, India \\
  \texttt{harsh777111raj@gmail.com} \\
  \And
  Domenic Rosati \\
  scite.ai \\
  Brooklyn, NY, USA \\
  \texttt{dom@scite.ai} \\
  \And
  Subhabrata Majumdar \\
  Trustworthy ML Initiative \\
  Seattle, WA, USA \\
  \texttt{zoom.subha@gmail.com}
}

\begin{document}

\maketitle

\begin{abstract}
While large pretrained language models (PLMs) demonstrate incredible fluency and performance on many natural language tasks, recent work 
has shown that well-performing PLMs are very sensitive to what prompts are feed into them. Even when prompts are semantically identical, language models may give very different answers
. When considering safe and trustworthy deployments of PLMs we would like their outputs to be consistent under prompts that mean the same thing or convey the same intent. While some work has looked into how state-of-the-art PLMs address this need, they have been limited to only evaluating lexical equality of single- or multi-word answers and do not address consistency of generative text sequences. In order to understand consistency of PLMs under text generation settings, we develop a measure of semantic consistency that allows the comparison of open-ended text outputs. We implement several versions of this consistency metric to evaluate the performance of a number of PLMs on paraphrased versions of questions in the TruthfulQA dataset, we find that our proposed metrics are considerably more consistent than traditional metrics embodying lexical consistency, and also correlate with human evaluation of output consistency to a higher degree.
\end{abstract}

\section{Introduction}
As large pretrained language models (PLMs) get adopted more and more into next-generation automated workflows for natural language-based tasks, concerns about the safety and reliability of such models have also come in prominence \citep{Weidinger}. In Natural Language Generation (NLG), it is reasonable to expect a reliable PLM that is prompted with semantically equivalent prompts to return semantically equivalent outputs. This property is called {\it consistency}. Without this property, \textit{we cannot ensure the safety of PLMs} since we won't be certain about the outputs of models under inputs that semantically equivalent to seen examples. However, the degree to which current approaches to NLG are consistent and how to measure their consistency are still unclear. Addressing this gap, we introduce a novel framework for measuring consistency of NLG that generalizes previous approaches measuring consistency evaluation of single-token outputs (such as \citep{elazar_measuring_2021}), to measuring consistency of generated text sequences. 

Consistency measurement was introduced in the LAMA probe to understand PLMs as knowledge bases \citep{petroni}. \citet{elazar_back_2021} extended this and developed the ParaRel dataset 
to evaluate the consistency of masked language models by understanding the tokens they would predict for masked tuples. Taking this direction of research forward, \citet{fierro_factual_2022} extended their methods to a multilingual, multi-token setting. \citet{jang_accurate} developed a novel framework for understanding consistency on fine-tuned models on sentence similarity tasks. \citet{zhou} developed an approach based on using multiple prompts to specify single tasks to improve consistency metrics by more than 10\% across diverse data and task settings. Finally, \citet{newman_p-adapters_2022} developed a robust method to accurately extract factual information from PLMs as measured by a consistency measure they introduce.




In this paper, we present a novel measure of consistency for NLG that goes beyond the simple lexical measures proposed in past work, encoding realistic variation across prompts that semantically carry the same meaning but are lexically dissimilar. We empirically evaluate the effectiveness of this metric for consistency evaluation using multiple measures of semantic equivalence, PLM architectures, and answer generation/decoding methods. We find that (1) PLMs tend to generate {\it less accurate} answers as their size increases (the so-called inverse scaling phenomenon, see \citet{lin-etal-2022-truthfulqa}), (2) PLMs tend to become {\it more consistent} with size. (3), consistency and accuracy are not correlated and seem to reflect orthogonal properties, and (4) \textit{the biggest contributor to inconsistency is the decoding method}. Greedy approaches to decoding are much more consistent than a sampling-based approach.

\section{Method}

\citet{elazar_measuring_2021} first formalized consistency analysis of PLMs. The goal of consistency in PLMs is that prompts which are paraphrases of each other should produce the equivalent outputs to be considered consistent. Given a set of $n$ semantically similar prompts: $X = \{ x_1, \ldots, x_n \}$, with generated outputs $Y = \{ y_1, \ldots, y_n \}$, the consistency measure of \citet{elazar_measuring_2021} is defined as
\begin{align}\label{eqn:consistency_old}
    \text{Cons}_{lex} (Y) = \frac{1}{n(n-1)}
    \sum_{i,j=1, i \neq j}^{n} \mathbb{I} (y_i = y_j).
\end{align}

A key limitation of the work of \citet{elazar_measuring_2021} and its follow-up studies \citep{fierro_factual_2022, jiang_how_2020,zhou, newman_p-adapters_2022} is that their experimental consistency evaluations of PLMs were limited to settings where outputs are considered consistent if they are {\it exactly equal} to each other, or have a certain amount of token overlap. This means \emph{we cannot use the same measure to evaluate freeform natural language outputs of language models}. Furthermore, the constraint of exact equality (or multi-token equality \citep{kassner-etal-2021-multilingual}) is not realistic where we can have consistent answers that are not lexically identical but are in fact semantically identical.

We extend the consistency measure in Eq.~\eqref{eqn:consistency_old} by relaxing the exact equality constraint and replacing it with an agreement function that can identify semantic agreement between two outputs such as whether they are paraphrases or whether there is an entailment relationship between them.

\begin{align}\label{eqn:consistency_new}
    \text{Cons}_{sem} (Y) = \frac{1}{n(n-1)}
    \sum_{i,j=1, i \neq j}^{n} f (y_i, y_j).
\end{align}

Eq.~\eqref{eqn:consistency_new} describes a generalization of the consistency metric in Eq.~\eqref{eqn:consistency_old} originally proposed in \citet{elazar_measuring_2021}, replacing token equality of outputs with a {\it semantic equivalence function} $f(\cdot)$ to produce a consistency score in the range $[0,1]$. Note that we can easily recover the original metric of \citet{elazar_measuring_2021} by setting the token equality indicator as $f(\cdot)$: $f(y_i, y_j) \equiv \mathbb I(y_i = y_j)$ . A key advantage of this generalization is {\it sequential composition}: if we have access to semantic equivalence agreement functions across diverse domains such as text, image, and audio generation, then we can use the same framework to evaluate consistency across multimodal generative tasks as well.

In this paper, we explore four such agreement functions: semantic similarity using BERTScore \citep[BERTs]{bertscore}, semantic equivalence using paraphrase detection (PP, using the approach in Appendix \ref{paraphrase-model-details}), and entailment (Entail) and contradiction (Contra) using a natural language inference model (DeBERTa v2 xxlarge trained on mnli). For comparison purposes, we also implement rouge1 (R1-C) and named-entity overlap (NER) as heuristic measures of token overlap.

\section{Experimental Setup}

To evaluate semantic consistency under the framework proposed above, we take the generated text responses of PLMs on the TruthfulQA benchmark dataset~\citep{lin-etal-2022-truthfulqa}, enriched with several paraphrases for each of the questions. We chose TruthfulQA because it has a series of metrics and baselines for evaluating freeform text generation. In order to have access to a diverse set of paraphrases for the questions in TruthfulQA, we utilize three methods of question paraphrase generation: DocT5Query \citep{Cheriton2019FromDT}, QC \citep{bandel-etal-2022-quality}, and GPT-3 \citep{gpt3} with few-shot prompts for generating paraphrases (using the method found in \citet{novelty}). This results in a total of 8,956 paraphrased questions.

Since having consistent answers requires that the question paraphrases actually ask the same thing, we train a paraphrase detection model on PAWS~\citep{zhang-etal-2019-paws} (see Appendix \ref{paraphrase-model-details} for details), and filter out all generated paraphrases that fail to be detected as paraphrase after passing through this model using a two-step process. First, we rank paraphrases by their output probabilities from the PAWS model of being a paraphrase, and select the top 6 paraphrases. Second, we manually remove question paraphrases which we do not consider to be actual paraphrases of the original question. These steps result in a final set of 3,962 paraphrased questions covering 817 original questions (see Appendix \ref{question-consistency} for results on unfiltered questions and further analysis).

To understand the effects of model parameter size on consistency, we prompt a series of OPT models \citep{opt} (from 125M to 2.7B parameters) with the question paraphrases and an answer prompt resulting in a set of answers for each question. We generate answers using greedy decoding and nucleus sampling \citep{curious_case_2019} procedures in order to understand how sampling-based decoding might effect consistency (inspired by \citet{factuality_enhanced} who found a dramatic effect of decoding method on factuality). Finally we run these set of answers through both the accuracy metrics for freeform text generation outlined in TruthfulQA \citep{lin-etal-2022-truthfulqa} and the semantic consistency metrics proposed above. We further evaluate a random sample of 100 questions with GPT-3 text-davinci-002 \citep{gpt3,instruct} to understand how the computed consistency values compare with those of a model that has a much larger number (175B) of parameters.

To validate our semantic consistency measurement framework and chosen agreement functions, we ran a human study with three volunteer participants who labeled a random sample of 100 questions with 903 answer pairs resulting from paraphrases of the 100 sampled questions. Participants were instructed to label answer pairs
as consistent if they considered the two answers as semantically equivalent and inconsistent otherwise. The inter-annotator agreement has a Fleiss' $\kappa$ value of 0.84, indicating strong agreement between participants. We use these human scores and perform a correlation study with our proposed semantic consistency scores and the TruthfulQA accuracy scores to understand what metrics might be most reflective of what humans understand as consistency. Since we do not have access to the Judge model or human evaluation scores from the TruthfulQA paper, we did not compare those scores with our approach.

\section{Results}
\label{headings}

\begin{table*}[t]
\caption{Accuracy and consistency of OPT and GPT-3 models on TruthfulQA}
  \label{consitency-accuracy-table}
  \centering
\begin{tabular}{@{}llllllllll@{}}
\toprule
Metric type & Acc &  & Consis &  &  &  &  &  &  \\ 
Model & R1-A & BLEURT & PP & PP+acc & BERTs & Entail & Contra & R1-C & NER \\ \midrule
Greedy &  &  &  &  &  &  &  &  &  \\ \midrule
OPT-125M & 40.0 & 50.0 & 27.7 & 21.4 & 90.7 & 26.1 & 25.7 & 12.0 & 12.2 \\
OPT-350M & 41.6 & 50.3 & 17.1 & 17.7 & 88.8 & 18.0 & 23.1 & 12.6 & 13.5 \\
OPT-1.3B & 36.5 & 49.2 & 30.6 & 25.2 & 89.5 & 23.6 & 29.4 & 12.7 & 11.9 \\
OPT-2.7B & 35.2 & 43.6 & 37.6 & 32.6 & 90.3 & 27.4 & 28.6 & 11.6 & 11.1 \\
GPT-3 & \textbf{59.0} & \textbf{62.6} & \textbf{62.2} & \textbf{71.5} & \textbf{92.3} & \textbf{42.5} & \textbf{11.3} & \textbf{30.4} & \textbf{19.8} \\ \midrule
Sampled &  &  &  &  &  &  &  &  &  \\ \midrule
OPT-125M & 41.9 & 50.7 & 6.2 & 2.5 & 86.1 & 5.2 & 36.7 & 0.2 & 4.7 \\
OPT-350M & 43.2 & 50.7 & 4.6 & 2.8 & 85.6 & 4.4 & 31.0 & 0.2 & 3.9 \\
OPT-1.3B & 40.1 & 50.4 & 11.3 & 6.0 & 85.6 & 5.7 & 36.0 & 0.3 & 3.7 \\
OPT-2.7B & 40.0 & 48.9 & 14.2 & 9.2 & 85.8 & 6.6 & 36.7 & 0.3 & 3.2 \\
GPT-3 & \textbf{56.3} & \textbf{60.5} & \textbf{54.0} & \textbf{66.0} & \textbf{90.8} & \textbf{32.0} & \textbf{14.1} & \textbf{19.9} & \textbf{14.2} \\ \bottomrule
\end{tabular}
\end{table*}

Table \ref{consitency-accuracy-table} outlines a number of key findings. \textbf{First}, we are able to reproduce the finding that as OPT models get larger in parameter size they tend to do worse at providing truthful answers in TruthfulQA \citep{lin-etal-2022-truthfulqa}, as measured by rouge1 (R1-A) and BLEURT \citep{sellam2020bleurt}. \textbf{Second}, there is a general trend of models became more consistent with parameter size especially in the case of paraphrase and entailment which correlated most with human judgements of consistency. This is also the case for answers that are both consistent under paraphrase and accurate (PP+acc). \textbf{Third}, while nucleus sampling is more accurate than greedy sampling for the OPT models, greedy sampling tends to be much more consistent on all consistency measures. Finally lexical measures of consistency (R1-C, NER) and BERTScore are relatively uninformative measures.

Table \ref{consistency-correlation-table} outlines results from the correlation analysis between each consistency metric and human scores. Another important finding here is that {\it accuracy measures do not correlate with consistency measures}. The two accuracy measures show moderate positive correlation with one another ($\rho=0.48$). However, consistency measures tend to not correlate with one another, except in the case of PP, entailment, and BERTscore measures where there are moderate positive correlations. Human scores show the strongest correlation with entailment ($\rho$=0.70), PP ($\rho$=0.52), and BERTScore ($\rho$=0.54), but have weak positive correlation with NER ($\rho$=0.15) and R1-c ($\rho$=0.32)
These observations indicate that our semantic consistency scores do indeed measure the property of consistency as understood by humans, and do a better job than lexical measures based on token similarity.

\begin{table*}[t]
\caption{Correlation of consistency scores (Spearman $\rho$) with accuracy scores as well as human consistency annotations. Scored on outputs from OPT-2.7B with greedy decoding.}
  \label{consistency-correlation-table}
  \centering
\begin{tabular}{@{}llllllllll@{}}
\toprule
 & Human & R1-C & NER & BERTs & PP & Entail & Contra & R1-A & BLEURT \\ \midrule
Human & 1.00 & 0.32 & 0.15 & 0.54 & 0.52 & 0.70 & -0.28 & -0.10 & -0.10 \\
R1-C & & 1.00 & 0.27 & 0.23 & -0.10 & 0.15 & -0.08 & 0.10 & -0.11 \\
NER & & & 1.00 & -0.05 & -0.07 & 0.01 & -0.05 & -0.05 & -0.06 \\
BERTs & & & & 1.00 & 0.66 & 0.78 & -0.29 & -0.13 & -0.22 \\
PP & & & & & 1.00 & 0.76 & -0.48 & -0.15 & -0.09 \\
Entail & & & & & & 1.00 & -0.49 & -0.25 & -0.22 \\
Contra & & & & & & & 1.00 & 0.05 & 0.11 \\
R1-A & & & & & & & & 1.00 & 0.48 \\
BLEURT & & & & & & & & & 1.00 \\\bottomrule
\end{tabular}
\end{table*}

Given the above results, we recommend that entailment and paraphrase-based agreement functions should be further investigated as consistency measures. Lexical approaches do not correlate well with human judgement. While BERTScore does have a similar correlation with human judgement as paraphrase, this is due to the fact that what BERTScore really measures is semantic {\it similarity}. Two outputs can still be semantically \textit{similar} while being inconsistent. This is not the case with the paraphrase-based approach. Additionally, Table~\ref{consitency-accuracy-table} shows that BERTScore is relatively uninformative for analyzing models as they scale. 

\section{Conclusion}

In this work, we presented a novel framework for evaluating consistency of NLG that correlates well with human judgements when using paraphrase or entailment as agreement functions. While we have demonstrated that our semantic consistency framework works well for measuring consistency in NLG, future work should continue to validate our proposal across other types of text generation tasks such as dialogue or table-to-text generation. Additionally, a key limitation of our work is that any error in the agreement function will be reflected as error in the consistency score. Finally, since \citet{lin-etal-2022-truthfulqa} has shown an inverse scaling on TruthfulQA, future work should follow up with datasets where scaling laws hold to understand how consistency looks on those.

Many of the above findings are intuitive or in line with previous results. \citet{lin-etal-2022-truthfulqa} found similar inverse scaling on TruthfulQA. While \citet{webson_prompt-based_2022} has shown that PLMs have poor understanding of prompts, and prompts are better understood by larger models. If larger models understand their prompts more, it makes sense that they would be more consistent than smaller ones. \citet{factuality_enhanced} found sampling-based decoding was much less factual then greedy decoding with the explanation that sampling introduces randomness causing factual errors. Along these lines it is plausible that sampling-based decoding methods could be introducing inconsistency by randomness during decoding. This is worth further investigation in future studies.

Less intuitive is that consistency and accuracy do not correlate. Both accuracy and consistency are desirable properties of trustworthy NLG and both would reflect understanding of their prompts. Yet, since there doesn't appear to be a strong relationship between accuracy and consistency, simply measuring accuracy is not enough to guarantee reliable PLMs. Future work could extend and validate our framework by using it as the basis for measurable improvements of consistency in NLG settings.

\section*{Reproducibility}

The GitHub repository \url{https://github.com/harshraj172/Measuring-Reliability-of-LLMs} contains our final dataset of TruthfulQA paraphrases.

\bibliographystyle{plainnat}
\bibliography{main}

\begin{thebibliography}{26}
\providecommand{\natexlab}[1]{#1}
\providecommand{\url}[1]{\texttt{#1}}
\expandafter\ifx\csname urlstyle\endcsname\relax
  \providecommand{\doi}[1]{doi: #1}\else
  \providecommand{\doi}{doi: \begingroup \urlstyle{rm}\Url}\fi

\bibitem[Bandel et~al.(2022)Bandel, Aharonov, Shmueli-Scheuer,
  et~al.]{bandel-etal-2022-quality}
Elron Bandel, Ranit Aharonov, Michal Shmueli-Scheuer, et~al.
\newblock Quality controlled paraphrase generation.
\newblock In \emph{Proceedings of the 60th Annual Meeting of the Association
  for Computational Linguistics (Volume 1: Long Papers)}, pages 596--609,
  Dublin, Ireland, May 2022. Association for Computational Linguistics.
\newblock \doi{10.18653/v1/2022.acl-long.45}.
\newblock URL \url{https://aclanthology.org/2022.acl-long.45}.

\bibitem[Brown et~al.(2020)Brown, Mann, Ryder, et~al.]{gpt3}
Tom~B. Brown, Benjamin Mann, Nick Ryder, et~al.
\newblock Language models are few-shot learners, 2020.
\newblock URL \url{https://arxiv.org/abs/2005.14165}.

\bibitem[Cheriton(2019)]{Cheriton2019FromDT}
David~R. Cheriton.
\newblock From doc2query to doctttttquery.
\newblock 2019.

\bibitem[Chowdhury et~al.(2022)Chowdhury, Zhuang, and Wang]{novelty}
Jishnu~Ray Chowdhury, Yong Zhuang, and Shuyi Wang.
\newblock Novelty controlled paraphrase generation with retrieval augmented
  conditional prompt tuning, 2022.
\newblock URL \url{https://arxiv.org/abs/2202.00535}.

\bibitem[Elazar et~al.(2021{\natexlab{a}})Elazar, Kassner, Ravfogel,
  et~al.]{elazar_measuring_2021}
Yanai Elazar, Nora Kassner, Shauli Ravfogel, et~al.
\newblock Measuring and {Improving} {Consistency} in {Pretrained} {Language}
  {Models}.
\newblock \emph{Transactions of the Association for Computational Linguistics},
  2021{\natexlab{a}}.
\newblock \doi{10.1162/tacl_a_00410}.

\bibitem[Elazar et~al.(2021{\natexlab{b}})Elazar, Zhang, Goldberg, and
  Roth]{elazar_back_2021}
Yanai Elazar, Hongming Zhang, Yoav Goldberg, and Dan Roth.
\newblock Back to {Square} {One}: {Artifact} {Detection}, {Training} and
  {Commonsense} {Disentanglement} in the {Winograd} {Schema}, October
  2021{\natexlab{b}}.
\newblock URL \url{http://arxiv.org/abs/2104.08161}.
\newblock arXiv:2104.08161 [cs].

\bibitem[Fierro and Søgaard(2022)]{fierro_factual_2022}
Constanza Fierro and Anders Søgaard.
\newblock Factual {Consistency} of {Multilingual} {Pretrained} {Language}
  {Models}.
\newblock In \emph{{FINDINGS}}, 2022.
\newblock \doi{10.48550/arXiv.2203.11552}.

\bibitem[He et~al.(2020)He, Liu, Gao, and Chen]{deberta}
Pengcheng He, Xiaodong Liu, Jianfeng Gao, and Weizhu Chen.
\newblock Deberta: Decoding-enhanced bert with disentangled attention, 2020.
\newblock URL \url{https://arxiv.org/abs/2006.03654}.

\bibitem[He et~al.(2021)He, Gao, and Chen]{debertav3}
Pengcheng He, Jianfeng Gao, and Weizhu Chen.
\newblock Debertav3: Improving deberta using electra-style pre-training with
  gradient-disentangled embedding sharing, 2021.
\newblock URL \url{https://arxiv.org/abs/2111.09543}.

\bibitem[Holtzman et~al.(2019)Holtzman, Buys, Du, et~al.]{curious_case_2019}
Ari Holtzman, Jan Buys, Li~Du, et~al.
\newblock The curious case of neural text degeneration, 2019.
\newblock URL \url{https://arxiv.org/abs/1904.09751}.

\bibitem[Jang et~al.(2021)Jang, Kwon, and Lukasiewicz]{jang_accurate}
Myeongjun Jang, Deuk~Sin Kwon, and Thomas Lukasiewicz.
\newblock Accurate, yet inconsistent? consistency analysis on language
  understanding models.
\newblock \emph{CoRR}, abs/2108.06665, 2021.
\newblock URL \url{https://arxiv.org/abs/2108.06665}.

\bibitem[Jiang et~al.(2020)Jiang, Xu, Araki, and Neubig]{jiang_how_2020}
Zhengbao Jiang, Frank~F. Xu, Jun Araki, and Graham Neubig.
\newblock How {Can} {We} {Know} {What} {Language} {Models} {Know}?, May 2020.
\newblock URL \url{http://arxiv.org/abs/1911.12543}.
\newblock arXiv:1911.12543 [cs].

\bibitem[Kassner et~al.(2021)Kassner, Dufter, and
  Sch{\"u}tze]{kassner-etal-2021-multilingual}
Nora Kassner, Philipp Dufter, and Hinrich Sch{\"u}tze.
\newblock Multilingual {LAMA}: Investigating knowledge in multilingual
  pretrained language models.
\newblock In \emph{Proceedings of the 16th Conference of the European Chapter
  of the Association for Computational Linguistics: Main Volume}, pages
  3250--3258, Online, April 2021. Association for Computational Linguistics.
\newblock \doi{10.18653/v1/2021.eacl-main.284}.
\newblock URL \url{https://aclanthology.org/2021.eacl-main.284}.

\bibitem[Lee et~al.(2022)Lee, Ping, Xu, et~al.]{factuality_enhanced}
Nayeon Lee, Wei Ping, Peng Xu, et~al.
\newblock Factuality enhanced language models for open-ended text generation,
  2022.
\newblock URL \url{https://arxiv.org/abs/2206.04624}.

\bibitem[Lin et~al.(2022)Lin, Hilton, and Evans]{lin-etal-2022-truthfulqa}
Stephanie Lin, Jacob Hilton, and Owain Evans.
\newblock {T}ruthful{QA}: Measuring how models mimic human falsehoods.
\newblock In \emph{Proceedings of the 60th Annual Meeting of the Association
  for Computational Linguistics (Volume 1: Long Papers)}, pages 3214--3252,
  Dublin, Ireland, May 2022. Association for Computational Linguistics.
\newblock \doi{10.18653/v1/2022.acl-long.229}.
\newblock URL \url{https://aclanthology.org/2022.acl-long.229}.

\bibitem[Newman et~al.(2022)Newman, Choubey, and
  Rajani]{newman_p-adapters_2022}
Benjamin Newman, Prafulla~Kumar Choubey, and Nazneen Rajani.
\newblock P-{Adapters}: {Robustly} {Extracting} {Factual} {Information} from
  {Language} {Models} with {Diverse} {Prompts}, April 2022.
\newblock URL \url{http://arxiv.org/abs/2110.07280}.
\newblock arXiv:2110.07280 [cs].

\bibitem[Ouyang et~al.(2022)Ouyang, Wu, Jiang, et~al.]{instruct}
Long Ouyang, Jeff Wu, Xu~Jiang, et~al.
\newblock Training language models to follow instructions with human feedback,
  2022.
\newblock URL \url{https://arxiv.org/abs/2203.02155}.

\bibitem[Petroni et~al.(2019)Petroni, Rockt{\"{a}}schel, Riedel,
  et~al.]{petroni}
Fabio Petroni, Tim Rockt{\"{a}}schel, Sebastian Riedel, et~al.
\newblock Language models as knowledge bases?
\newblock In Kentaro Inui, Jing Jiang, Vincent Ng, and Xiaojun Wan, editors,
  \emph{Proceedings of the 2019 Conference on Empirical Methods in Natural
  Language Processing and the 9th International Joint Conference on Natural
  Language Processing, {EMNLP-IJCNLP} 2019, Hong Kong, China, November 3-7,
  2019}, pages 2463--2473. Association for Computational Linguistics, 2019.
\newblock \doi{10.18653/v1/D19-1250}.
\newblock URL \url{https://doi.org/10.18653/v1/D19-1250}.

\bibitem[Sellam et~al.(2020)Sellam, Das, and Parikh]{sellam2020bleurt}
Thibault Sellam, Dipanjan Das, and Ankur~P Parikh.
\newblock Bleurt: Learning robust metrics for text generation.
\newblock In \emph{Proceedings of ACL}, 2020.

\bibitem[Webson and Pavlick(2022)]{webson_prompt-based_2022}
Albert Webson and Ellie Pavlick.
\newblock Do {Prompt}-{Based} {Models} {Really} {Understand} the {Meaning} of
  {Their} {Prompts}?
\newblock \emph{NAACL}, 2022.
\newblock \doi{10.18653/v1/2022.naacl-main.167}.

\bibitem[Weidinger et~al.(2022)Weidinger, Uesato, Rauh, et~al.]{Weidinger}
Laura Weidinger, Jonathan Uesato, Maribeth Rauh, et~al.
\newblock Taxonomy of risks posed by language models.
\newblock In \emph{2022 ACM Conference on Fairness, Accountability, and
  Transparency}, FAccT '22, page 214–229, New York, NY, USA, 2022.
  Association for Computing Machinery.
\newblock ISBN 9781450393522.
\newblock \doi{10.1145/3531146.3533088}.
\newblock URL \url{https://doi.org/10.1145/3531146.3533088}.

\bibitem[Williams et~al.(2017)Williams, Nangia, and Bowman]{mnli}
Adina Williams, Nikita Nangia, and Samuel~R. Bowman.
\newblock A broad-coverage challenge corpus for sentence understanding through
  inference, 2017.
\newblock URL \url{https://arxiv.org/abs/1704.05426}.

\bibitem[Zhang et~al.(2022)Zhang, Roller, Goyal, et~al.]{opt}
Susan Zhang, Stephen Roller, Naman Goyal, et~al.
\newblock Opt: Open pre-trained transformer language models, 2022.
\newblock URL \url{https://arxiv.org/abs/2205.01068}.

\bibitem[Zhang et~al.(2019{\natexlab{a}})Zhang, Kishore, Wu, et~al.]{bertscore}
Tianyi Zhang, Varsha Kishore, Felix Wu, et~al.
\newblock Bertscore: Evaluating text generation with bert, 2019{\natexlab{a}}.
\newblock URL \url{https://arxiv.org/abs/1904.09675}.

\bibitem[Zhang et~al.(2019{\natexlab{b}})Zhang, Baldridge, and
  He]{zhang-etal-2019-paws}
Yuan Zhang, Jason Baldridge, and Luheng He.
\newblock {PAWS}: Paraphrase adversaries from word scrambling.
\newblock In \emph{Proceedings of the 2019 Conference of the North {A}merican
  Chapter of the Association for Computational Linguistics: Human Language
  Technologies, Volume 1 (Long and Short Papers)}, pages 1298--1308,
  Minneapolis, Minnesota, June 2019{\natexlab{b}}. Association for
  Computational Linguistics.
\newblock \doi{10.18653/v1/N19-1131}.
\newblock URL \url{https://aclanthology.org/N19-1131}.

\bibitem[Zhou et~al.(2022)Zhou, He, Ma, Berg-Kirkpatrick, and Neubig]{zhou}
Chunting Zhou, Junxian He, Xuezhe Ma, Taylor Berg-Kirkpatrick, and Graham
  Neubig.
\newblock Prompt consistency for zero-shot task generalization, 2022.

\end{thebibliography}

\section*{Appendix}
\appendix
\section{Paraphrase model details}
\label{paraphrase-model-details}

We finetuned a DeBERTa v3 \citep{debertav3} large model on PAWS \citep{zhang-etal-2019-paws} paraphrase dataset for paraphrase detection and used a threshold on 0.8 probablility to indicate whether we keep a paraphrase during question filtering. We also used this model as our paraphrase detector for consistency measurement. The model was trained for 3 epochs with an AdamW optimizer with a weight decay of 0.01, warmup steps of 50, batch size of 8, and learning rate of 6e-6.

We used a pretrained DeBERTa base model \citep{deberta} trained on MNLI \citep{mnli} for entailment and contradiction measures.

\section{Question consistency and answer consistency}
\label{question-consistency}

We can use our evaluation framework to understand the consistency of the prompts we used before and after human filtering. Both question sets were surprisingly consistent when using paraphrase as an agreement function with 93.34\% consistency of question paraphrases without human filtering and 92.42\% consistency of question paraphrases after automatic and human filtering. Since we would expect human filtering to have improved consistency, future work should follow up on what the discrepancy might be between filtering inconsistent results and measuring them.

To show what the accuracy and consistency results were when using the unfiltered set we present Table \ref{consitency-accuracy-table-unfiltered} below that summarizes results on the unfiltered set. Table \ref{consitency-accuracy-table-unfiltered} shows that they are almost the same results as Table \ref{consitency-accuracy-table} presented before with a downward accuracy trend, improved paraphrase consistency, and worse consistency with sampling-based decoding methods.

\begin{table}[]
\caption{Accuracy and consistency of OPT and GPT-3 models on TruthfulQA on unfiltered questions}
  \label{consitency-accuracy-table-unfiltered}
  \centering
\begin{tabular}{@{}llllllll@{}}
\toprule
 & Acc &  & Consis &  &  &  &  \\ 
model & R1-A & BLEURT & PP & PP+acc & BERTScore & R1-C & NER \\\midrule
greedy &  &  &  &  &  &  &  \\\midrule
opt-125m & 40.0 & 50.0 & 26.2 & 19.7 & 90.2 & 9.9 & 10.5 \\
opt-350m & 41.6 & 50.3 & 16.1 & 14.3 & 88.4 & 10.7 & 12.0 \\
opt-1.3b & 36.5 & 49.2 & 28.8 & 22.3 & 89.0 & 10.5 & 10.8 \\
opt-2.7b & 35.2 & 43.6 & 35.0 & 28.6 & 89.6 & 9.5 & 9.7 \\\midrule
sampled &  &  &  &  &  &  &  \\\midrule
opt-125m & 41.9 & 50.7 & 6.0 & 2.5 & 86.0 & 0.2 & 3.8 \\
opt-350m & 43.2 & 50.7 & 4.4 & 2.7 & 85.4 & 0.2 & 3.3 \\
opt-1.3b & 40.1 & 50.4 & 11.2 & 5.4 & 85.5 & 0.2 & 3.1 \\
opt-2.7b & 40.0 & 48.9 & 14.4 & 9.2 & 85.7 & 0.3 & 2.8 \\ \bottomrule
\end{tabular}
\end{table}

\end{document}